\documentclass[letterpaper, 10 pt, conference]{ieeeconf}

\IEEEoverridecommandlockouts
\usepackage{amsmath,amsfonts}
\usepackage{booktabs}
\usepackage{multirow}
\usepackage{graphicx}
\usepackage{makecell}
\usepackage{algorithmic}
\usepackage[ruled,noend]{algorithm2e} % For algorithm.
\usepackage{array}
\usepackage[caption=false,font=normalsize,labelfont=sf,textfont=sf]{subfig}
\usepackage{textcomp}
\usepackage{stfloats}
\usepackage{url}
\usepackage[table]{xcolor}
\usepackage{verbatim}
\def\BibTeX{{\rm B\kern-.05em{\sc i\kern-.025em b}\kern-.08em
    T\kern-.1667em\lower.7ex\hbox{E}\kern-.125emX}}
\usepackage{balance}

\usepackage[backref=page]{hyperref}
\hypersetup{
     colorlinks = true,
     linkcolor = green,
     anchorcolor = green,
     citecolor = green,
     filecolor = green,
     urlcolor = cyan
 }

\newcommand{\ours}[0]{Simify}
\newcommand{\task}[1]{\textit{#1}}
\newcommand{\method}[1]{\textsc{#1}}
\begin{document}

\title{\LARGE \bf
Test-Time Spatial Reasoning for Robot Manipulation\\Using Generative Real-to-Sim
}
\author{\authorblockN{Ivan Kapelyukh\textsuperscript{1,2}, 
Yafei Hu\textsuperscript{1}, 
Ran Gong\textsuperscript{1}, 
Brandon May\textsuperscript{1}, 
Tushar Kusnur\textsuperscript{1}, \\
Laura Herlant\textsuperscript{1}, 
Karl Schmeckpeper\textsuperscript{1}, 
Edward Johns\textsuperscript{2,\S}, 
Xiaohan Zhang\textsuperscript{1,\S}}
\authorblockA{\textsuperscript{1}Robotics and AI Institute \qquad \textsuperscript{2}Imperial College London \qquad
\textsuperscript{\S}Project Leads}}

\makeatletter
\twocolumn[{%
\renewcommand\twocolumn[1][]{#1}%
\maketitle
\begin{center}
    \vspace{-0.7cm}
    \includegraphics[width=0.93\linewidth]{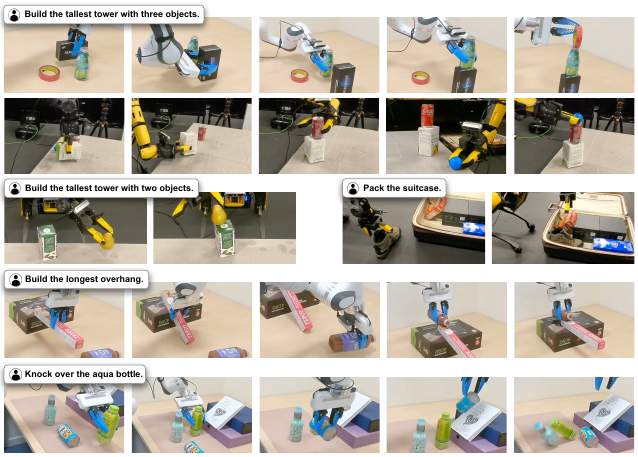}
    \def\@captype{figure}
    \caption{Zero-shot spatial and physical reasoning with \ours{}. From a single RGB-D view and task reward, the robot reconstructs the scene, searches for object arrangements in simulation, and then executes the rearrangement in the real world.}
    \label{fig:teaser}
\end{center}%
}]
\makeatother

\thispagestyle{empty}
\pagestyle{empty}

\begin{abstract}
Spatial reasoning is fundamental to general robot intelligence, as it enables robots to complete long-horizon tasks involving multi-object interaction.
We introduce \ours{}, a training-free, test-time framework that performs explicit spatial reasoning via massively parallel physics simulation.
From a single RGB-D image of a scene, \ours{} reconstructs simulation-ready assets leveraging 3D generative models and vision-language models.
Then given a task specified by a reward function (e.g., \textit{build the tallest tower}), \ours{} launches thousands of parallel rollouts in simulation and performs an evolutionary search to optimize object arrangements, typically converging within seconds.
We conduct quantitative experiments on real-robot hardware to demonstrate the ability of our framework to execute complex object rearrangement tasks end-to-end with previously unseen objects. 
Results show that our framework outperforms prior work on foundation models for spatial reasoning by effectively exploiting large-scale parallel simulation during inference, and also highlight the importance of complete and accurate geometry for successful sim-to-real transfer.

\end{abstract}

\section{Introduction}

Spatial reasoning refers to the ability to understand object geometry, spatial relationships, and feasible arrangements in 3D space in order to achieve a goal.
For robots, spatial reasoning must often be physically grounded: the robot must not only infer where objects can be placed, but also whether the resulting arrangement will remain stable, produce the desired contact interactions, and be executable in the real world.
Despite rapid progress in training spatial-aware Vision-Language Models (VLMs), replicating such physically grounded spatial reasoning in continuous 3D space remains a fundamental challenge for robots.

To endow robots with spatial reasoning capability, recent works have largely explored two directions: using pre-trained 3D-aware foundation models~(\cite{lerf2023, openeqa, 3dllm, voxposer, gu2024conceptgraphs}), or training end-to-end Vision-Language-Action (VLA) policies~(\cite{intelligence2025pi_05, gr00tn1_2025, molmoact2025}).
However, geometric accuracy and physical grounding remain major bottlenecks for these approaches: while they capture high-level semantic relationships, tasks like stacking and balancing (Fig.~\ref{fig:teaser}) require high precision and strict adherence to physical dynamics -- capabilities that implicit learned representations often fail to guarantee.
On the other hand, VLAs and other imitation learning approaches are developed by scaling data to solve a diverse set of such manipulation tasks. 
Despite the large amount of teleoperation data they require during training, these approaches tend not to discover new strategies on novel tasks at test time; instead, their generalization is mostly imitation of action patterns similar to those the robot has seen before \cite{demystifying-diffusion}. 

In this work, we focus on a complementary regime to policy learning: test-time spatial reasoning for novel object rearrangement. Rather than assuming task-specific demonstrations, offline training data of successful arrangements, known object models, or a learned goal-state predictor, we ask whether a robot can construct a physical model of the current scene and search for a solution directly at test time. This setting is especially important for object and position generalization: when previously unseen objects or new initial poses change the feasible placement order, orientation, or contact configuration, success may require reasoning about the specific geometry and physical interactions in the current scene. While imitation learning and RL are powerful when sufficient task distributions and training data are available, they are not the focus of this paper. Instead, our method is closer in spirit to test-time planning and TAMP-style reasoning, but targets open-world scenes where object models must be inferred from a single RGB-D observation.

As humans, we approach such problems through a process of deliberation: we can mentally rotate, place, and evaluate candidates before taking any actions, even when encountering the objects for the first time. 
Analogously, robots should also simulate and search for solutions at test time and plan accordingly. 
This paradigm shifts the burden from learning all geometry and physics inside one visuomotor policy to using external tools: \textit{a 3D physics simulator and a search-based optimization procedure for explicit reasoning.} 

\begin{figure*}[t]
    \centering
    \includegraphics[width=0.95\linewidth]{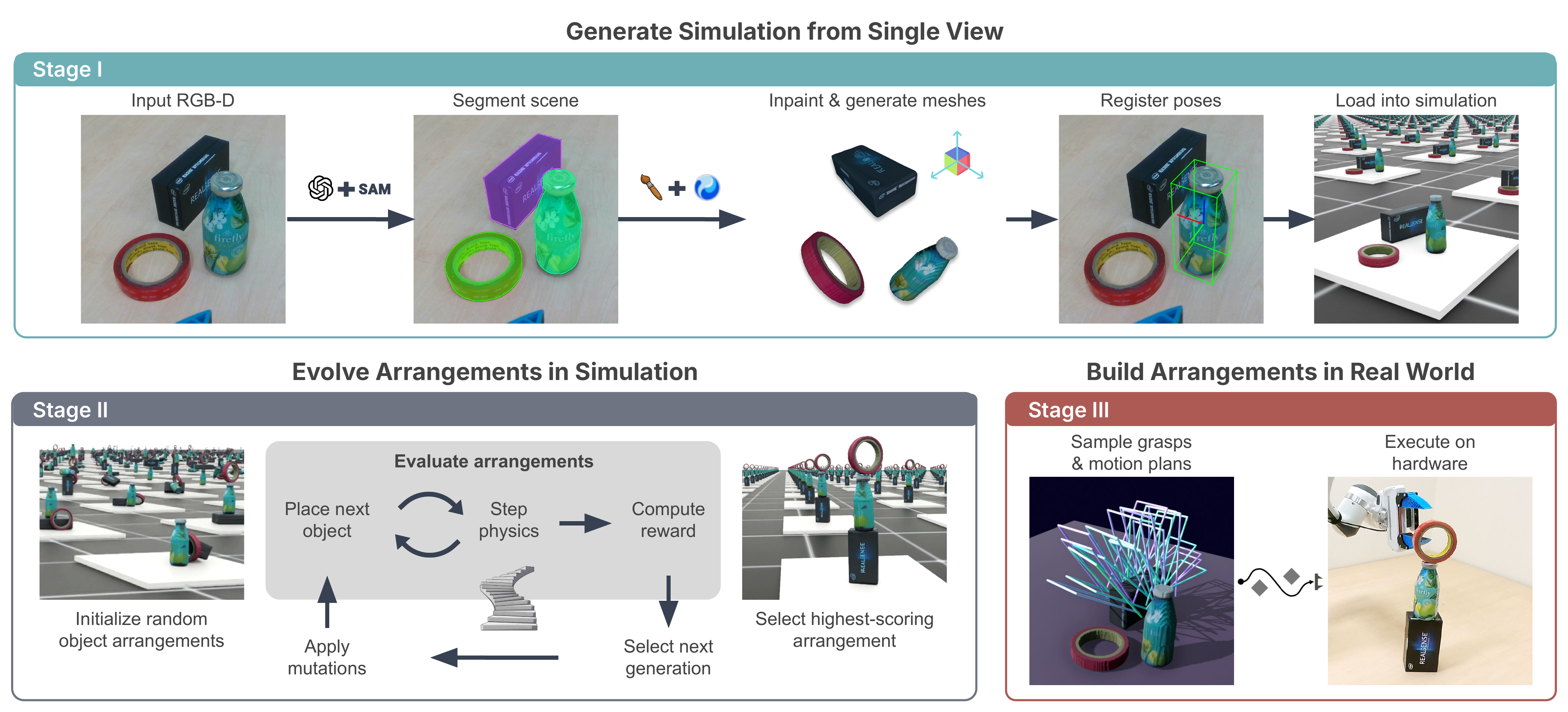}
    \vspace{-1em}
    \caption{Overview of \ours{}. (1) From a single RGB-D view, the system segments and completes objects, generates scaled meshes, estimates their poses, and builds a parallel physics simulation. (2) Evolutionary search jointly optimizes placement order and goal object poses, followed by noisy validation to select a robust, high-reward arrangement. (3) The robot tracks objects, samples grasps, and executes collision-free pick-and-place motions.}
    \label{fig:overview}
    \vspace{-1em}
\end{figure*}

We introduce \ours{}, a training-free framework for solving tasks that require spatial creativity (Fig.~\ref{fig:overview}). 
From a single RGB-D image, \ours{} reconstructs a simulation-ready digital twin of previously unseen environments using 3D generative priors and LLM/VLMs. 
Specifically, we leverage open-vocabulary segmentation and 2D inpainting models to resolve occlusions, feeding the resulting object images into an image-to-3D diffusion model to generate complete, scaled meshes. 
Once instantiated in a massively parallel physics simulator, \ours{} uses an evolutionary algorithm to jointly optimize the discrete placement sequence and continuous goal poses for each object. 
By evaluating thousands of noisy physical rollouts in parallel, the framework effectively searches for robust, executable arrangements that maximize the reward function given for the task. 
Finally, the optimal arrangement is transferred back to the real world and executed via a vision-based, collision-free motion planning pipeline.

We quantitatively evaluate \ours{} both in simulation and on real-robot hardware.
Results show the importance of complete and accurate geometry for downstream task success, and highlight how our framework can be applied to a range of spatial reasoning tasks.

\section{Related work}
\label{sec:citations}

\subsection{Generative Real-to-Sim}
Real-to-sim methods build a simulation of the scene observed by the robot, enabling the robot to reason, learn, or plan in a reconstructed physical environment before acting in the real world. Prior work has studied real-to-sim for estimating physical properties such as mass, friction, contact dynamics, and articulation~\cite{closing-loop,asid,contact-dynamics,real2sim2real-casting,urdformer}, as well as visual reconstruction using depth scans, NeRFs, or Gaussian Splatting~\cite{rialto,embodied-gaussians,vlm-keypoints}. However, these approaches often require known object models, dense multi-view scanning, manual setup, or multiple cameras, and they struggle to recover occluded object geometry from a single robot observation. In this work, we use recent image-to-3D generative models as part of our real-to-sim pipeline: given a single RGB-D view, the robot segments objects, completes occluded regions, generates full object meshes with a visual generative prior, scales and registers them to the observed scene, and instantiates them in physics simulation. By leveraging powerful vision priors, our framework minimizes human effort and enables the reconstruction of unobserved surfaces (e.g. the bottom of an object), which we show is important for accurate dynamics in simulation.

\subsection{Real-to-Sim in Robotics}
Several recent robotics systems have also begun to use image-to-3D models or real-to-sim techniques for manipulation, including grasp sampling, reconstruction of human demonstrations, policy evaluation, dataset generation, and rapid reinforcement learning in simulation~\cite{scenecomplete,video2policy,phystwin,robogen,gen2sim,robotwin,zerobot}. These works are complementary to ours, but they primarily use the generated simulation to train or evaluate policies, often assuming that the task goal or demonstration is already provided. For example, ZeroBot~\cite{zerobot} uses generative real-to-sim to train a state-based RL policy that manipulates a novel object to a given goal pose. In contrast, our work studies a different regime: test-time spatial reasoning, where the robot must infer the goal arrangement itself. To our knowledge, this is the first work to explore single-view generative real-to-sim for test-time physics-based planning to solve these long-horizon, multi-object spatial reasoning tasks, with no human demonstrations or training datasets of arrangements required.

\subsection{Object Rearrangement and TAMP}
Object rearrangement requires choosing both a discrete placement order and continuous goal poses, making it a natural setting for spatial reasoning~\cite{rearrangement}. Task and Motion Planning (TAMP) is a strong framework for such long-horizon manipulation problems, as it jointly reasons over symbolic action sequences and continuous geometric feasibility~\cite{tamp,diffusion-constraints}. Recent open-vocabulary TAMP systems further use VLMs or LLMs to ground language instructions, infer goals, or propose constraints for planning~\cite{llm-grop,zhang2024dkprompt}. We therefore compare against \method{LLM-GROP} as a strong zero-shot TAMP baseline.
Our setting differs from most prior rearrangement work in that success depends not only on geometric feasibility, but also on physical outcomes after execution, such as stability, balance, and dynamic object-object interaction. Traditional TAMP methods often assume known or reconstructed object models and hand-designed predicates or constraints, while learned rearrangement methods require training datasets of successful arrangements, thus requiring more human effort to apply to a new task~\cite{liu2022structformer,structdiffusion,diffusion-constraints}. In contrast, our method is training-free: from a single RGB-D observation and a task reward, it reconstructs the current scene and uses massively parallel physics simulation at test time to search over placement order and continuous object poses. This enables real-to-sim-to-real spatial reasoning with everyday objects, including tasks such as \task{Stacking}, \task{Cantilever}, and \task{Dominoes}.

\section{Problem Statement}
\label{sec:problem_statement}
We consider the challenging multi-object rearrangement setup, and we specifically tackle this problem in environments containing novel, everyday objects without relying on human demonstrations or object models known a priori.

\noindent \textbf{Sensory Input \& Task Specification:}
The robot is provided with an initial state of the environment, captured by a single-view RGB-D observation, denoted as $I$. 
The task specification $\mathcal{T}$ is provided either as a natural-language description or a reward function. 
In this work, we formalize the task as a scalar reward function $R_{\mathcal{T}}$\footnote{See Section~\ref{sec:experiments} for reward function definitions. We assume $R_{\mathcal{T}}$ is given; synthesizing rewards from language is left for future work~\cite{eureka, gong2025anytask}.} that evaluates the geometric and structural success of the post-execution scene.

\noindent \textbf{Action \& Solution:}
Solving the rearrangement task requires the robot to output a single ordered sequence of $N$ desired object goal poses, denoted as $\mathbf{G} = (G_1, G_2, \dots, G_N)$. The order of this sequence intrinsically defines the discrete placement sequence, while each $G_i$ specifies the continuous 6D target pose for that object $i$ in the world frame. 

We assume the robot sequentially executes these goal poses using a vision-based pick-and-place skill. 
While this skill could be an end-to-end visuomotor policy, in this work we implement it via a traditional pipeline integrating pose estimation, grasp sampling, and collision-free motion planning. More details can be found in Section~\ref{sec:methodology}.

\noindent \textbf{Dynamics \& Objective:}
Real-world physical execution is inherently non-deterministic. 
After a robot places an object at its goal pose $G_i$, it settles into a final pose $\tilde{G}_i$ due to physical interactions, ranging from quasi-static structural settling to dynamic multi-object collisions (e.g., \task{Dominoes}).
We denote this ``place-then-settle'' physical transition as a mapping $\mathcal{S}$, which induces the final settled poses based on the ordered sequence of object goal poses: $\tilde{G} = \mathcal{S}(\mathbf{G})$.

The overarching objective is to discover an executable rearrangement sequence $\mathbf{G}^{*}$ that maximizes the expected task reward on the fully settled scene:
\[
\mathbf{G}^{*}
=
\arg\max_{\mathbf{G}}
\mathbb{E}\!\left[R_{\mathcal{T}}(\tilde{\mathbf{G}})\right]
\quad
\text{s.t.}\quad
\tilde{\mathbf{G}}=\mathcal{S}(\mathbf{G}).
\]
We rely on simulated physics to estimate $\mathcal{S}$ prior to real-world execution. In this work, we focus on real2sim for geometry, and assume that physics parameters such as mass and friction are given, following prior work \cite{rialto}.

\section{Methodology}
In order to address this problem, we propose the {\ours} framework, an overview of which is shown in Fig. \ref{fig:overview}. First, our method builds a simulation of the scene using mesh generation (Section \ref{ss:real2sim}). Once in simulation, the poses of the objects are optimized using an evolutionary algorithm, in order to find an arrangement with a high task reward (Section \ref{ss:evolution}). After the goal pose for each object has been determined, the rearrangement is executed on the real robot using motion planning (Section \ref{ss:execution}).

\label{sec:methodology}

\subsection{Real-to-sim}\label{ss:real2sim}

The objective of this stage is to construct a simulation-ready mesh for each object. The robot captures a single-view RGB-D image $I$ using a wrist-mounted camera; we assume a given viewpoint and leave autonomous active vision to future work~\cite{dm-osvp}. To reconstruct the scene without task-specific instructions or predefined object names, we use bottom-up segmentation. SAM-2~\cite{sam2} first generates segment proposals, which may over-segment objects or include background regions. A VLM (GPT-4o in our experiments) lists the manipulable objects in $I$ and assigns the numbered SAM-2 segments to each object. This produces manipulation-relevant object masks with minimal human input, while allowing optional user instructions to modify the segmentation.

Because image-to-3D models such as Hunyuan3D-2~\cite{hunyuan3d2} are typically trained on unoccluded object views, we first complete each masked object crop using a 2D amodal inpainting model~\cite{gen3dsr} based on Stable Diffusion~\cite{ldm}. The model receives the object crop with other segments masked out and a VLM-generated text description, producing an occlusion-free image for mesh generation. Since the generated mesh lacks metric scale, we align it to the observed partial depth point cloud using ICP. We then register the scaled mesh to the scene with Foundation Pose, which estimates the initial object pose $T_{WO}$ from the RGB-D image and generated mesh. The resulting meshes and poses are loaded into Isaac Lab~\cite{isaaclab} for massively parallel GPU-accelerated simulation. Our pipeline also supports multiple views by tracking segments with SAM-2 and aggregating depth point clouds for scale estimation; however, we evaluate the single-view setting because it avoids additional scanning and matches the input regime of current image-to-3D models.

\subsection{Evolutionary Optimization in Simulation}\label{ss:evolution}

\begin{algorithm}[t]
\small

\caption{Evolutionary Optimization of Multi-Object Rearrangement}
\label{alg:evolve_arrangements}
\KwIn{$N$ object (meshes) $\{\mathbf{M}^{(1)}, \dots, \mathbf{M}^{(N)}\}$, Reward function $R_{\mathcal{T}}$, Population size $L$, Max generations $\gamma$, Validation rollouts $K$}
\KwOut{$\mathbf{G}^*$}

Initialize population $\mathcal{P}_0 = \{\mathbf{G}^{(1)}, \dots, \mathbf{G}^{(L)}\}$\;

\For{$g \gets 0$ \KwTo $\gamma-1$}{
  \ForEach{$\mathbf{G} \in \mathcal{P}_{g}$ \textbf{in parallel}}{
    Reset simulation\;
    \For{object $i \gets 1$ \KwTo $N$}{
      Add Gaussian noise: ${G}_i = G_i + \epsilon$\;
      Place object $i$ at ${G}_i$\;
      Step physics until stable;
    }
    Compute task reward $r$;
  }
  
  Rank population $\mathcal{P}_{g}$ by $r$\;
  Keep top $5\%$ of $\mathbf{G} \in \mathcal{P}_{g}$ unchanged for $\mathcal{P}_{g+1}$\;
  Sample the remaining via softmax over rewards\;
  Mutate goal poses and placement order\;
}

\vspace{1mm}
\textbf{Validation Stage:}\;
\ForEach{$\mathbf{G} \in \mathcal{P}_{\gamma}$ \textbf{in parallel}}{
    \For{$k \gets 1$ \KwTo $K$}{
        Execute $\mathbf{G}$ with $\epsilon$\;
        Obtain $\tilde{G}$ and compute reward $r_k = R_{\mathcal{T}}(\tilde{G})$\;
    }
    Compute expected reward: $\bar{r} = \frac{1}{K} \sum_{k=1}^{K} r_k$\;
}
\Return $\mathbf{G}^* = \arg\max_{\mathbf{G}} \bar{r}$\;
\end{algorithm}

\textbf{Objective.} Given reconstructed object meshes and a task reward, we optimize the placement order and 6D goal poses to maximize the final arrangement reward. We use an evolutionary algorithm (Algorithm~\ref{alg:evolve_arrangements}) because the search space is mixed discrete--continuous, multimodal, and well suited to massively parallel simulation.

\textbf{Evolution.} We initialize $L$ solutions with random placement orders and object poses (Fig.~\ref{fig:overview}). Each solution is evaluated in a parallel simulation by sequentially placing the objects, allowing the scene to settle after each placement, and computing the final task reward. The top $5\%$ are retained as elites, while the remaining solutions are sampled according to a reward-based softmax and mutated in both pose and placement order. This process favors high-reward solutions while maintaining population diversity.

\textbf{Validation.} To account for reconstruction and execution errors, Gaussian placement noise is applied during optimization. After the final generation, each candidate is evaluated over multiple noisy rollouts, and the solution with the highest expected reward is selected for real-world execution. Increasing the noise scale favors robust solutions, while decreasing it permits more precise but higher-risk arrangements.

\subsection{Real-Robot Execution}\label{ss:execution}

Once we have a goal pose for each object, and the order in which to place objects, we execute the rearrangement on the real robot. First, we observe the scene from the original camera view, and use Foundation Pose to estimate the latest pose of each object. Then, for each object, we sample grasps using the geometry of the generated mesh. Specifically, we uniformly sample across the triangles of the mesh to determine a contact point \cite{uniform-mesh-sampling}, and then sample a gripper orientation based on the surface normal at that point. Since we have complete meshes for each object, we can eliminate grasps which are in collision with the environment. We then sample collision-free motion plans for picking and placing the object from its initial pose into its goal pose. 

We use the cuRobo library \cite{curobo} allowing us to perform thousands of collision checks in parallel using CUDA acceleration. For tasks such as stacking, the placement must be precise, and our pose estimation may have error as it is tracking against a generated, imperfect mesh. Therefore, for the final portion of the placement trajectory, we use the force sensor on the end-effector to move the object downward until a sufficient increase in force is detected, indicating that contact has been made with the supporting surface, and the object can be released. This closes the loop on the pre-planned trajectory to avoid dropping the object early.

\section{Experiments}
\label{sec:experiments}

\subsection{Research Questions}

We explore the following research questions: (Q1) How effective is our framework at spatial reasoning, compared with the web-scale prior of vision-language models? (Q2) How important is the 3D prior in real2sim for downstream task success? (Q3) Do solutions from our simulation-based method transfer to the real world? (Q4) How do our design decisions affect performance? (Q5) How accurate and complete are the generated meshes?

\subsection{Experimental Setup}

\noindent \textbf{Robot system.} To evaluate our framework end-to-end on a real robot system, we use a 7-DoF Franka Panda robot equipped with a compliant gripper and an RGB-D wrist camera (RealSense D435), which captures an image of the scene from a given pose, for input to our real2sim pipeline.

\noindent \textbf{Evaluation tasks.} We evaluate our method on a set of challenging spatial reasoning tasks using everyday household objects (Fig. \ref{fig:teaser}). Note that the robot must perform these tasks on these novel objects without human demonstrations or a training dataset of arrangements.

\noindent\textbf{Stacking}: the robot must stack the objects (a camera box, a glass bottle, and a roll of tape) into the tallest tower possible. The $(x, y)$ position of the stack is fixed, and the robot must optimize the order in which the objects are placed, as well as their $z$ positions (i.e. height above the table), and their orientations about the $x$ and $y$ axes (where the $+y$ direction faces left from the robot's base frame). This task requires careful long-horizon reasoning about structural stability, spatial creativity in reorienting objects, accurate geometric reconstruction of surfaces, and precise placement during execution. For this task, the reward function is defined to be the height of the tallest point in the scene above the table. The reward is always awarded at the end of the episode, after all objects have been placed and the scene has settled.

\noindent\textbf{Cantilever}: there is a large support box on the table, which cannot be moved. The robot can move two objects: a heavy smoothie bottle and a baking paper box. The task is to build the longest structure possible extending out from the support box over the table, without the structure falling and touching the table surface. The $x$ position of the structure is fixed. The robot must optimize the order in which the objects are placed, as well as their $z$ and $y$ positions and orientations about the $z$ axis, to build a structure which maximizes the length of the cantilever while maintaining balance. In order to make this task feasible for a unimanual system, we choose a large support box which is weighed down and has high-friction tape applied to its edge. The reward function is defined as the length of the structure built out from the support box, and if any object touches the table the reward is set to 0.

\noindent\textbf{Dominoes}: the robot must knock over the target object (the bottom-heavy, blue tea bottle), using the other objects as ``dominoes''. Specifically, the robot must choose the $x$, $y$, and $z$ positions of a tall green bottle (within the scene bounds of 70 cm $\times$ 10 cm $\times$ 70 cm from the scene origin), and a heavy tin can, as well as their placement order. There is a static pile of objects including a book which cannot be moved by the robot, but can be used as a ramp by placing other objects on top of it. This task requires spatial creativity, and reasoning about dynamic object-object interactions. The reward function is such that the further the target object has been moved from its original pose after the scene has settled, the higher the reward.

\subsection{Evaluating Spatial Reasoning in Simulation}
\begin{figure}[t]
    \centering
    \includegraphics[width=0.8\linewidth]{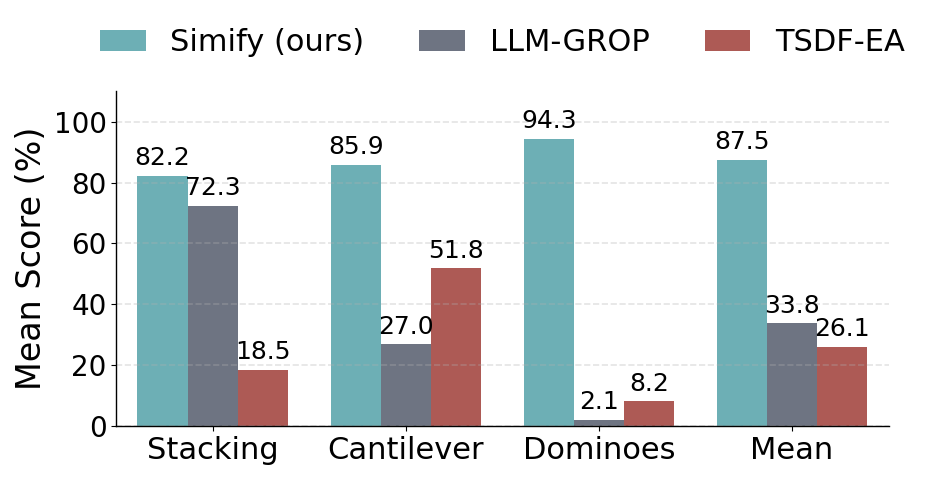}
    \caption{Normalized task scores for \ours{}, \method{LLM-GROP}, and \method{TSDF-EA}. \method{LLM-GROP} uses a VLM for goal-pose reasoning, while \method{TSDF-EA} uses partial TSDF geometry. \ours{} consistently finds higher-scoring arrangements.}
    \label{fig:sim_eval_results_comparison}
\end{figure}

In this section we address research questions Q1 and Q2, by comparing our method against the following baselines:

\noindent \textbf{1) \method{LLM-GROP}}~\cite{llm-grop}: a recent TAMP framework that uses large language models to infer goal poses for object rearrangement tasks. \method{LLM-GROP} is a particularly relevant baseline for our setting because it performs zero-shot goal-pose reasoning without requiring task-specific demonstrations or policy training. It therefore represents a strong test-time planning approach, and directly addresses the question of whether high-level semantic reasoning from an LLM is sufficient for solving our spatial rearrangement tasks. To ensure that the baseline has access to the same input information as our method, the prompt to the language model (GPT-5) includes the reward function, any axis constraints, the image of the scene annotated with the object and scene coordinate frames, and a text scene description including object names, masses, and 3D bounding boxes (obtained using our real2sim pipeline). The LLM is resampled until a parsable output with object order and 3D poses is produced.

\noindent \textbf{2) \method{TSDF-EA}}~\cite{tsdf}: The \method{TSDF-EA} baseline (Truncated Signed Distance Field -- Evolutionary Algorithm) ablates our use of a 3D prior in the real2sim pipeline: to obtain the mesh for each object, this baseline uses the input RGB-D image to reconstruct the visible geometry via TSDF~\cite{tsdf}, and then uses the same evolutionary algorithm as our method once in simulation.

\begin{figure}
    \centering
    \includegraphics[width=\linewidth]{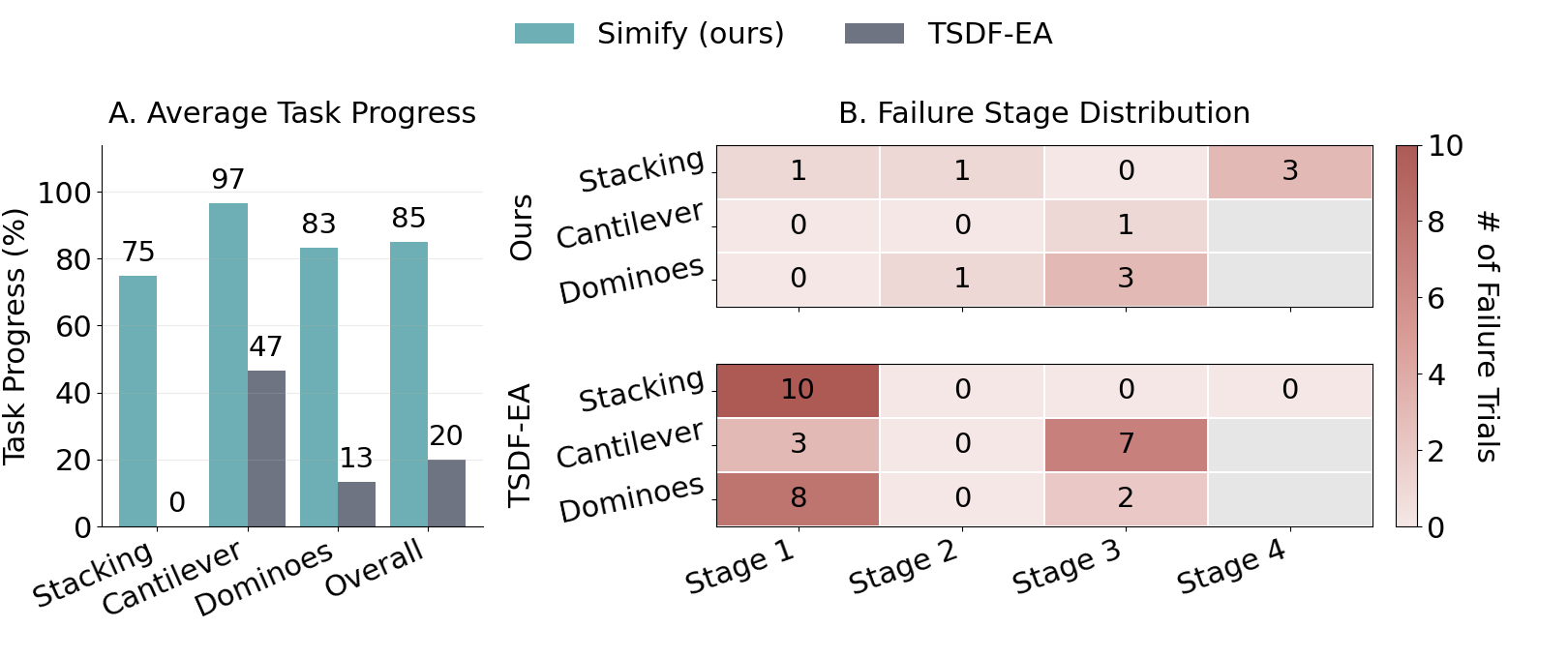}
\caption{Real-robot results over 10 trials per task. (A) Mean task progress. (B) Failure-stage distribution, where Stage 1 is planning and later stages are sequential object placements. \ours{} outperforms \method{TSDF-EA} and more often reaches later execution stages.}
    \label{fig:real_splitted}
\end{figure}

We evaluate each method's spatial reasoning solution in simulation. We conduct 10 repeats for each task and for each method, capturing a new image of the initial scene (varying the object poses by a small amount, since they must all fit in the same single view), and then running the full reconstruction and arrangement optimization pipelines, and saving the goal poses output from each method. We then load this solution into an evaluation simulation. Each object is placed into the goal pose determined by the method, with placement noise added, and after the scene settles, we assign a score using the reward function for that task. Note that to evaluate each solution, we use the same mesh in simulation as the mesh used by the method to find the solution during optimization: sim2real transfer is studied in the next section. As we are using simulation, we conduct 2048 evaluation repeats per solution. To allow aggregation across tasks, we rescale the reward into a task score (\%) by normalising between a minimum and maximum reward for each task.

The results are shown in Fig. \ref{fig:sim_eval_results_comparison}. Our method outperforms \method{TSDF-EA} on all tasks, showing the importance of using 3D priors in real2sim for generating complete meshes, particularly for tasks where accurately reconstructing unobserved surfaces is important for task success: for example, although the bottom surface of the tin can is in contact with the table and thus cannot be directly observed, our use of image-to-3D models allows this to still be reconstructed accurately, enabling the can to roll down the ramp correctly in simulation.  The \method{LLM-GROP} baseline performs well on \task{Stacking}, but is not able to reliably complete the others, whereas our method achieves a higher score on each task. While the VLM understands the high-level semantics of a task such as \task{Stacking}, it is less capable of determining the continuous pose at which to place each object, which is necessary for tasks requiring precise reasoning about object-object interaction, such as \task{Cantilever} or \task{Dominoes}. Our use of a physics simulator as a spatial reasoning engine during deployment enables our method to iteratively refine the poses and complete more precise tasks.

\subsection{Robustness of Real-World Execution}

We address Q3 by evaluating \ours{} end-to-end on real-robot hardware including execution. We compare \method{} with \method{TSDF-EA} to study the effect of mesh completion on sim2real transfer. For all methods, we terminate a rollout as a planning failure if the selected goal poses achieve a simulation success rate below a fixed threshold, indicating that the arrangement is unstable or unlikely to transfer.

\begin{figure}
    \centering
    \includegraphics[width=\linewidth]{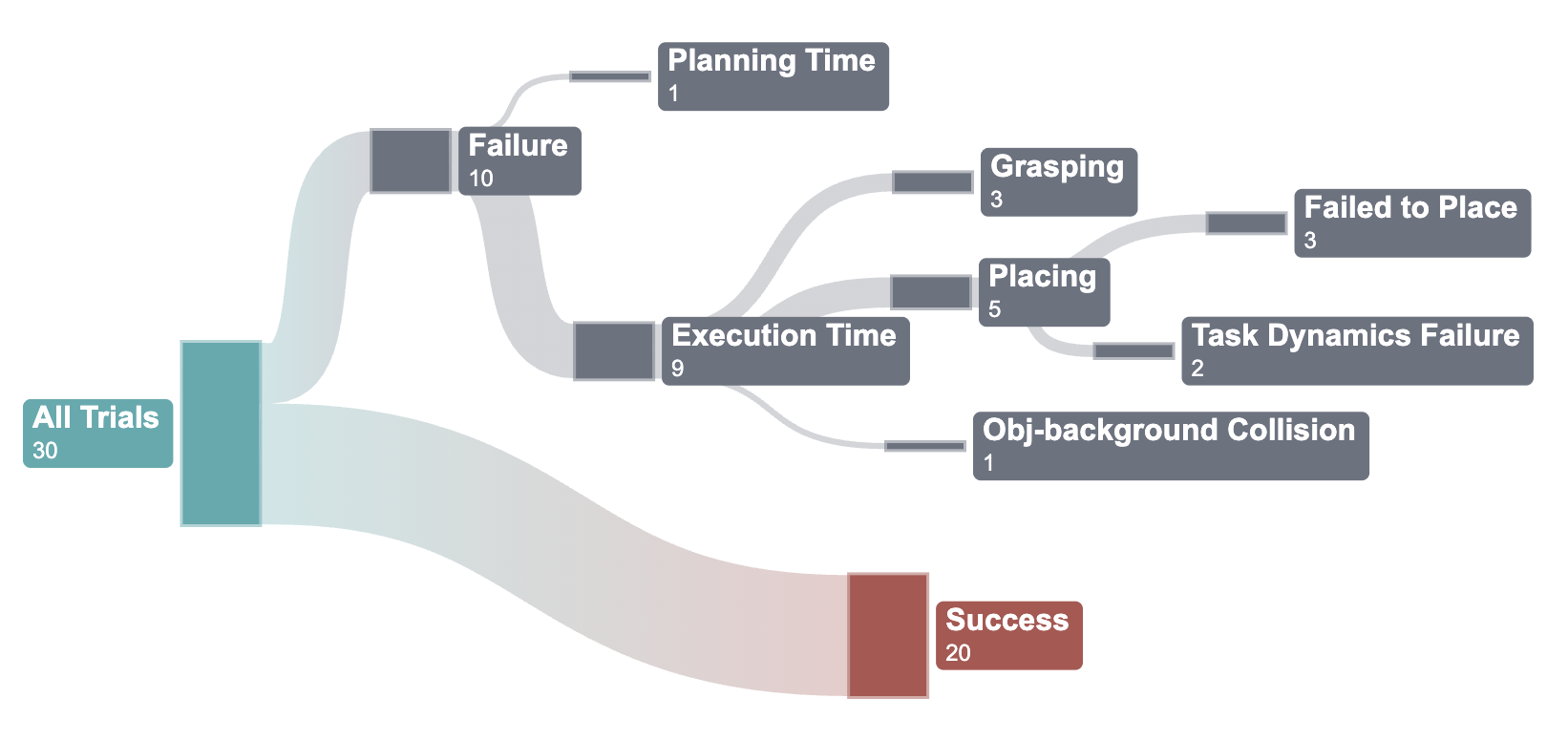}
    \caption{
    Failure breakdown for \ours{} over 30 real-world trials.
    }
    \label{fig:failure}
\end{figure}

Fig.~\ref{fig:real_splitted} summarizes real-world task progress and failure stages. \ours{} achieves substantially higher average task progress than \method{TSDF-EA} across all three tasks.
The failure-stage distribution further shows that \method{TSDF-EA} often fails early, especially in \task{Stacking}, where all 10 trials fail at Stage 1 (i.e., the planning stage). This indicates that partial geometry from the visible point cloud is often insufficient for discovering stable and executable arrangements. In contrast, the complete meshes generated by \ours{} lead to fewer planning failures and allow the robot to make progress to later execution stages.
In addition, Fig.~\ref{fig:failure} provides a detailed breakdown of the outcomes for \ours{}. 

\begin{table}[htbp]
    \centering
    \caption{Task score (\%) results for an ablation study, exploring the design choices in our evolutionary algorithm.}
    \label{tab:ablations}
    \begin{tabular}{lcccc}
    \toprule
    Method & \task{Stacking} & \task{Cantilever} & \task{Dominoes} & Mean \\
    \midrule
    MeshGen-Rand & 29.8 & 0.0 & 1.1 & 10.3 \\
    Simify-NoVal & 62.6 & 45.6 & 83.8 & 64.0 \\
    Simify (ours) & \textbf{82.2} & \textbf{85.9} & \textbf{94.3} & \textbf{87.5} \\
    \bottomrule
    \end{tabular}
\end{table}

\subsection{Ablating Design Choices for Evolutionary Optimization}

\begin{figure}[t]
    \centering
    \includegraphics[width=0.8\linewidth]{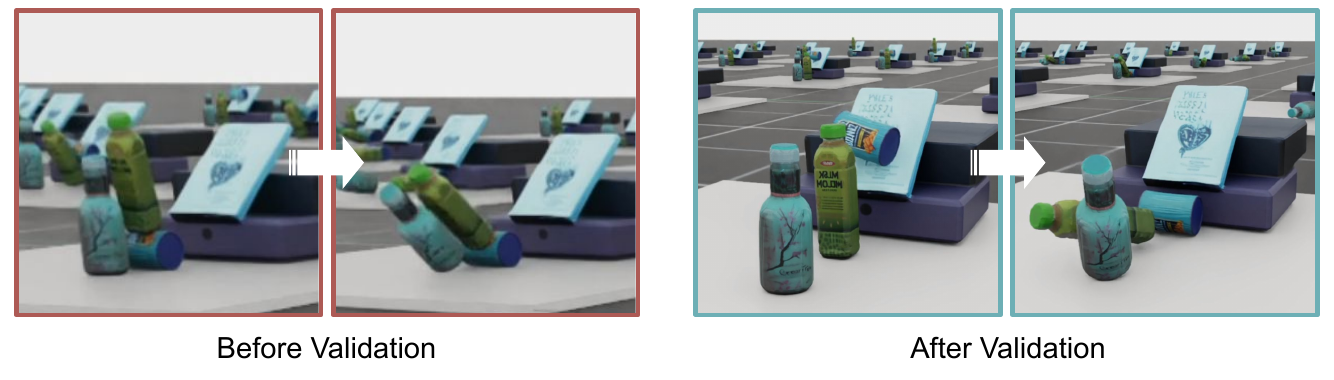}
    \caption{\task{Dominoes} solutions. \textbf{Left:} A high-risk strategy that places the green bottle on the can to tip it toward the target; it occasionally succeeds but is sensitive to placement noise. \textbf{Right:} The selected robust strategy, which places the bottle between the objects and rolls the can down the ramp. Validation favors the solution with higher expected reward.}
    \label{fig:qual_results_dominoes}
\end{figure}

To answer Q4, we ablate the main design choices in our evolutionary optimization algorithm. Table~\ref{tab:ablations} compares \ours{} against two variants. \method{MeshGen-Rand} uses the same generated meshes from our real2sim pipeline, but replaces our evolutionary optimization and validation algorithm with randomly sampled object poses and placement orders. Its low performance shows that complete geometry alone is not sufficient: the tasks require explicit spatial reasoning over the rearrangement sequence and continuous object poses. \method{Simify-NoVal} removes the final validation stage, where candidate solutions are evaluated across multiple noisy rollouts to estimate expected reward under placement error. 
This variant performs better than random search, but substantially worse than \ours{}, showing that selecting solutions based only on their best optimization score can produce arrangements that are brittle to execution noise (see Fig. \ref{fig:qual_results_dominoes} for an illustrative example). In contrast, our validation stage favors robust arrangements that remain successful under perturbations.

\begin{figure}
    \centering
    \includegraphics[width=\linewidth]{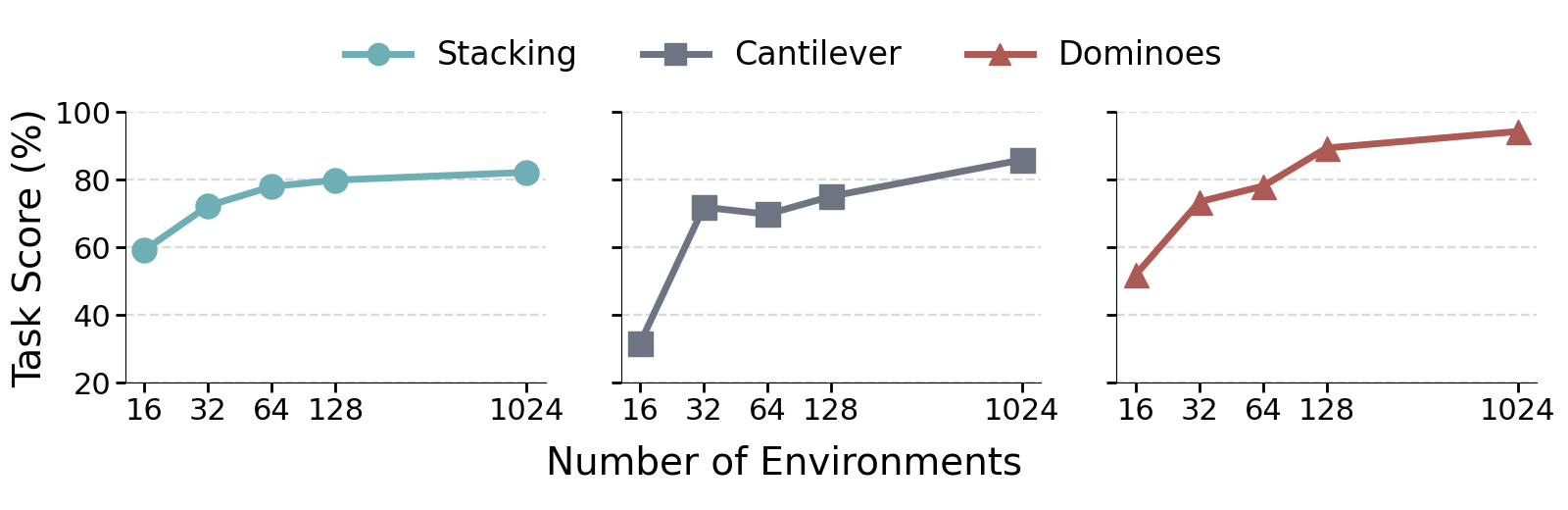}
    \vspace{-2em}
    \caption{Effect of simulation parallelism on task score. More environments improve performance on all tasks, showing that \ours{} benefits from additional test-time compute.}
    \label{fig:scaling}
\end{figure}

We further study how performance scales with the number of parallel simulation environments. Fig.~\ref{fig:scaling} shows that increasing the number of environments improves task score across all three tasks. This trend is especially clear for \task{Cantilever} and \task{Dominoes}, where larger populations help the optimizer discover more precise and physically robust arrangements. These results indicate that \ours{} can effectively use parallel compute during test-time reasoning: more simulation environments provide broader exploration of the mixed discrete-continuous solution space and lead to higher-quality solutions. With our default setting of 1024 environments, optimization takes only 73 seconds on average, making the approach practical for real-robot deployment.

\subsection{The Effect of Mesh Quality on Task Performance}
\begin{table}[h]
    \centering
\caption{Surface reconstruction quality of generated and partial TSDF meshes, measured against ground-truth YCB meshes.}
    \label{tab:object_metrics}
    
    \resizebox{1.0\columnwidth}{!}{%
        \begin{tabular}{lllcccc}
            \toprule
            & & & \makecell{\textbf{Chamfer} \\ \textbf{Completeness} $\downarrow$}
              & \makecell{\textbf{Chamfer} \\ \textbf{Accuracy} $\downarrow$}
              & \textbf{F1-Score} $\uparrow$
              & \makecell{\textbf{Normal} \\ \textbf{Consistency} $\uparrow$} \\
            \midrule
            % Lower Occlusion
            \multirow{10}{*}{\makecell[l]{\textbf{Low}\\\textbf{Occ.}}} & \multirow{5}{*}{\makecell[l]{TSDF\\\cite{tsdf}}} & Bottle & 0.015 & 0.003 & 0.687 & 0.752 \\
             &  & Bowl & 0.009 & 0.003 & 0.801 & 0.787 \\
             &  & Box & 0.026 & 0.002 & 0.596 & 0.791 \\
             &  & Pitcher & 0.022 & 0.002 & 0.642 & 0.728 \\
            & & \cellcolor{gray!20}\textit{Mean} & \cellcolor{gray!20}0.018 & \cellcolor{gray!20}\textbf{0.002} & \cellcolor{gray!20}0.682 & \cellcolor{gray!20}0.765 \\
            \cmidrule{2-7}
             & \multirow{5}{*}{\makecell[l]{Mesh\\Gen.\\(ours)}} & Bottle & 0.004 & 0.004 & 0.939 & 0.917 \\
             &  & Bowl & 0.002 & 0.003 & 0.996 & 0.904 \\
             &  & Box & 0.003 & 0.003 & 0.975 & 0.945 \\
             &  & Pitcher & 0.005 & 0.007 & 0.797 & 0.840 \\
            & & \cellcolor{gray!20}\textit{Mean} & \cellcolor{gray!20}\textbf{0.004} & \cellcolor{gray!20}0.004 & \cellcolor{gray!20}\textbf{0.927} & \cellcolor{gray!20}\textbf{0.902} \\
            \midrule
            % Higher Occlusion
            \multirow{10}{*}{\makecell[l]{\textbf{High}\\\textbf{Occ.}}} & \multirow{5}{*}{\makecell[l]{TSDF\\\cite{tsdf}}} & Bottle & 0.016 & 0.002 & 0.656 & 0.756 \\
             &  & Bowl & 0.018 & 0.003 & 0.602 & 0.723 \\
             &  & Box & 0.036 & 0.003 & 0.457 & 0.749 \\
             &  & Pitcher & 0.024 & 0.002 & 0.600 & 0.722 \\
            & & \cellcolor{gray!20}\textit{Mean} & \cellcolor{gray!20}0.024 & \cellcolor{gray!20}\textbf{0.002} & \cellcolor{gray!20}0.579 & \cellcolor{gray!20}0.738 \\
            \cmidrule{2-7}
             & \multirow{5}{*}{\makecell[l]{Mesh\\Gen.\\(ours)}} & Bottle & 0.004 & 0.004 & 0.986 & 0.907 \\
             &  & Bowl & 0.002 & 0.004 & 0.967 & 0.872 \\
             &  & Box & 0.009 & 0.010 & 0.556 & 0.778 \\
             &  & Pitcher & 0.005 & 0.008 & 0.790 & 0.838 \\
            & & \cellcolor{gray!20}\textit{Mean} & \cellcolor{gray!20}\textbf{0.005} & \cellcolor{gray!20}0.007 & \cellcolor{gray!20}\textbf{0.825} & \cellcolor{gray!20}\textbf{0.849} \\
            
            \bottomrule
        \end{tabular}%
    }
\end{table}

To answer Q5, we compare our mesh generation pipeline against the TSDF partial meshes using reconstruction metrics. We run our real2sim pipeline on scenes composed of YCB objects \cite{ycb}, for which ground-truth meshes are available. We align the estimated mesh against the ground-truth mesh using ICP. Chamfer Completeness measures the Euclidean distance from each sampled point on the ground-truth mesh to the nearest point on the estimated mesh, while Chamfer Accuracy measures the distance from each point on the estimated mesh to the nearest on the ground-truth mesh. The F1-Score is the harmonic mean of precision and recall, calculated as the proportion of points satisfying a $0.01$ m distance threshold. Normal Consistency uses the dot-product between the mesh surface normals. 

Results in TABLE~\ref{tab:object_metrics} show that while the depth-based TSDF meshes are marginally more accurate on visible surfaces, the generated meshes are significantly more complete due to the powerful 3D prior used, and so have a lower Chamfer Completeness distance, as well as a higher overall F1-Score and Normal Consistency. Our mesh generation pipeline is also more robust to occlusions than the baseline which does not use visual priors. While we leave an in-depth comparison of image-to-3D models on standard benchmarks to prior work \cite{image-3d-limits}, this shows that for our spatial reasoning problem setting, geometric completeness is important for downstream task performance and sim2real transfer.

\section{Conclusions, Limitations, and Future Work}
We present \ours, a framework for solving spatial reasoning tasks involving multi-object interaction, by first building a simulation of the scene using the powerful visual prior of image-to-3D models, and then discovering object arrangements which solve the task through evolutionary search using massively parallel physics simulation at inference time. Results show the importance of complete and accurate geometry for downstream task success, and demonstrate how our framework can perform multi-stage spatial reasoning tasks with novel objects and no training arrangements required.
\begin{figure}[t]
    \centering
    \includegraphics[width=0.8\linewidth]{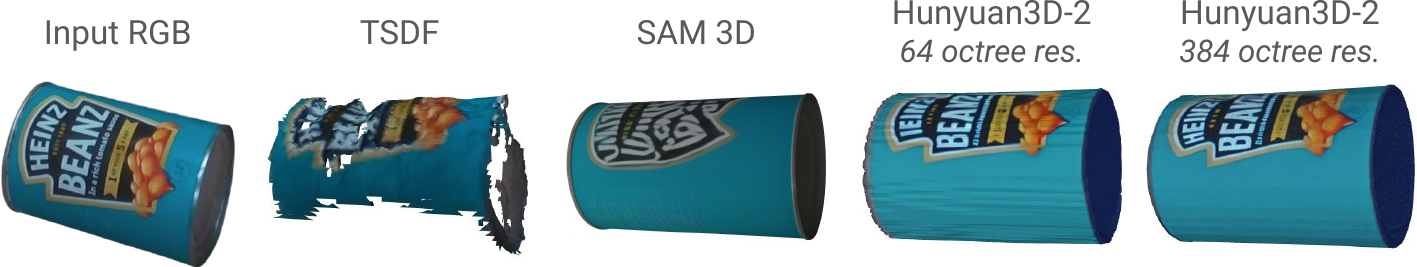}
    \caption{Qualitative comparison of partial TSDF, SAM3D, and Hunyuan3D-2 meshes at different octree resolutions. We use Hunyuan3D-2 at resolution 64 to balance reconstruction speed and accuracy; the modular pipeline also supports alternative image-to-3D models.}
    \label{fig:meshgen_qual_results}
\end{figure}

In this work, we focus on the role of geometry in real2sim for spatial reasoning, and assume that physics parameters such as mass and friction are given. We assume uniform density when the center of mass is set, and find that our method is sensitive to these parameters (in particular, in \task{Cantilever}). Automatically inferring these is an active area of research \cite{asid,real2sim2real-casting}. Future work can incorporate these techniques, and could also extend our framework beyond rigid objects. Another limitation of our method is that in order to simplify and accelerate simulation optimization, we place objects directly into their placement poses without a simulated robot. While we do use heuristics to filter out kinematically infeasible solutions (e.g. not sampling rotations directly away from the robot), the rearrangement may not always be feasible to execute directly. 
While most recent image-to-3D models use single-view input (Fig. \ref{fig:meshgen_qual_results}), and our framework inherits their limitations including precision \cite{image-3d-limits}, future image-to-3D models which take advantage of multi-view input can be swapped into our framework. Finally, combining our test-time simulation-based method with a VLM for high-level reasoning \cite{simpact} is a promising future work direction: the VLM could consider object semantics (our simulation only considers geometry), set the constraints for the optimization, and propose initial guesses to accelerate planning, particularly for larger scenes.

\balance
\bibliographystyle{IEEEtran}
\bibliography{simify}

\end{document}